\documentclass[conference]{IEEEtran}

\usepackage[T1]{fontenc}
\usepackage[utf8]{inputenc}
\usepackage[english]{babel}
\usepackage{csquotes}
\MakeOuterQuote{"}
\usepackage{xcolor}

\usepackage{amsmath}
\usepackage{amssymb}
\usepackage{mathtools}
\usepackage{commath}
\usepackage{bm}
\usepackage{dsfont}
\usepackage{bbm}

\usepackage{graphicx}
\usepackage{wrapfig}
\usepackage{caption,subcaption}
\usepackage{enumitem}
\usepackage{booktabs}
\usepackage{hhline}
\usepackage{color,soul}

\usepackage{hyperref}
\usepackage{cleveref}
\usepackage{tabularx}

\newcolumntype{C}{>{\centering\arraybackslash}X}
\usepackage{adjustbox}
\usepackage{svg}
\usepackage[backend=bibtex,sorting=none]{biblatex}
\AtEveryBibitem{
    \ifentrytype{artwork}{}{
    	\ifentrytype{online}{}{
        	\clearfield{url}
        	\clearfield{urldate}
        	\clearfield{urlyear}
        	\clearfield{urlmonth}
        }
    }
}

\usepackage{array} 
\usepackage{pifont} 

\DeclareMathOperator*{\argmax}{arg\,max}

\definecolor{UPblue}{rgb}{0, 0.353, 0.631}
\definecolor{UPred}{rgb}{0.898, 0.192, 0.22}
\definecolor{UPgray}{rgb}{0.6, 0.624, 0.62}
\definecolor{UPgreen}{rgb}{0, 0.376, 0.224}
\definecolor{UPorange}{rgb}{0.949, 0.58, 0}

\newcommand{\etal}{\textit{et al}.}
\newcommand{\ie}{{i}.{e}.~}

\newcommand{\secrefrange}[2]{\Cref{#1} to~\Cref{#2}}

\graphicspath{{images/}}

\def\BibTeX{{\rm B\kern-.05em{\sc i\kern-.025em b}\kern-.08em
    T\kern-.1667em\lower.7ex\hbox{E}\kern-.125emX}}

\begin{document}

\title{\title{Towards an Improved Metric for Evaluating Disentangled Representations}}

\author{
    \IEEEauthorblockN{Sahib Julka, Yashu Wang, Michael Granitzer}
    \IEEEauthorblockA{\textit{University of Passau} \\
     Germany \\
    firstname.lastname@uni-passau.de}
}

\maketitle
\thispagestyle{plain}
\pagestyle{plain}
\begin{abstract}
Disentangled representation learning plays a pivotal role in making representations controllable, interpretable and transferable. Despite its significance in the domain, the quest for reliable and consistent quantitative disentanglement metric remains a major challenge. This stems from the utilisation of diverse metrics measuring different properties and the potential bias introduced by their design. Our work undertakes a comprehensive examination of existing popular disentanglement evaluation metrics, comparing them in terms of measuring aspects of disentanglement (viz. Modularity, Compactness, and Explicitness), detecting the factor-code relationship, and describing the degree of disentanglement.  We propose a new framework for quantifying disentanglement, introducing a metric entitled \emph{EDI}, that leverages the intuitive concept of \emph{exclusivity} and improved factor-code relationship to minimize ad-hoc decisions. An in-depth analysis reveals that EDI measures essential properties while offering more stability than existing metrics, advocating for its adoption as a standardised approach.

\end{abstract}


\section{Introduction}

The learning of effective representations is crucial for enhancing the performance of downstream tasks in various domains. As defined by Bengio~\etal~\cite{bengio_representation_2014}, representation learning transforms observations into a format that captures the essence of data's inherent patterns and structures. An ideal representation should exhibit five key characteristics: (a) \textit{Disentanglement}, ensuring separate encoding of interpretable factors; (b) \textit{Informativeness}, capturing the diversity of data; (c) \textit{Invariance}, maintaining stability across changes in unrelated dimensions; (d) \textit{Compactness}, summarising essential information efficiently; and (e) \textit{Transferability}, facilitating application across different contexts. These attributes collectively enhance the model's interpretability, efficiency, and adaptability across tasks and domains.

While the literature does not present a unified theory of disentanglement, the consensus leans towards the principle that generative factors of variation ought to be individually encapsulated within distinct latent codes in the representation space. For instance, in an image dataset of human faces, an effective disentangled representation would feature separate dimensions for each identifiable attribute, such as face size, hairstyle, eye colour, and facial expression, among others.

The concept of \textit{modularity} or factor independence stemming from Independent factor analysis~\cite{attias1999independent} supports a commonly accepted view on disentanglement \cite{bengio_representation_2014, eastwood_dci_2022, ridgeway_fstatistic_2018}. This notion assumes no causal dependencies among the encoded dimensions, suggesting that in an ideally modularised representation, each generative factor is represented by a unique code or an independent subset of codes. As a result, modifying a specific code or subset within the representation space should ideally influence only its corresponding generative factor, leaving others unchanged.

An alternative perspective on disentanglement, rooted in the concept of compactness, posits that a generative factor should be represented by no more than a single code. This conceptualisation of disentanglement, emphasising the compactness and singularity of representation for each generative factor, has been adopted as a defining criterion by studies such as~\cite{kumar_sap_2018, chen_tcvae_2019}, and is also referred to as \textit{completeness}~\cite{eastwood_dci_2022}. Regardless of debates surrounding the desirability of compactness~\cite{ridgeway_fstatistic_2018, carbonneau_metrics_2022}, these concepts, along with modularity, have been embraced as part of a more comprehensive yet stringent framework for understanding disentanglement~\cite{eastwood_dci_2022, ridgeway_fstatistic_2018}. This integrated approach, which considers modularity, compactness, and explicitness, also known correspondingly as disentanglement, completeness, informativeness, has gained traction in more recent scholarly reviews on the topic~\cite{sepliarskaia_how_2021, carbonneau_metrics_2022}. Accordingly, a metric designed to quantify modularity and compactness should also assess informativeness~\ie, the extent to which latent codes encapsulate information about generative factors. When the ground truth factors of variation are identifiable, this informativeness transforms into explicitness, denoting the comprehensive representation of all recognised factors~\cite{higgins_definition_2018}.

Despite significant advancements in disentangling latent spaces via deep latent variable models \cite{higgins_beta-vae_2017, kim_factorising_2019, chen_tcvae_2019}, the literature still lacks a reliable and unified metric for evaluation. Traditionally, evaluation has been qualitative, relying on visual interpolation. The quantitative metrics that are available vary across the literature, and it has been demonstrated that the outcomes of these metrics do not consistently align with the findings from qualitative studies of disentangled representations~\cite{abdi_preliminary_2019, sepliarskaia_how_2021, locatello_challenging_2019}. Due to the variability in outcomes, a common measurement criterion has yet to be established. Furthermore, we observed that most existing metrics fail in certain scenarios and cannot be considered reliable across all settings, even when there is general agreement among them. Through an extensive analysis of the metrics, we identify these shortcomings and propose a new metric that is theoretically sound, reflects the desired properties better and is experimentally more robust. 

Concretely, in this work, our contributions can be summarised as:

\begin{itemize}

    \item We analyse the popular quantitative disentanglement metrics, identify their theoretical underpinnings, elaborate on the differences, and demonstrate their performance under various simulated conditions. 
    \item Based on the identified shortcomings, we propose a new metric called EDI, built on the novel principle of exclusivity. We show this metric performs better compared to the existing metrics on tests measuring calibration, non-linearity and robustness under noise, while being computationally efficient.
    \item We present a high-quality open-source codebase for reproducing our results and further research in this direction: \url{https://github.com/julka01/InnVariant}.
\end{itemize}

\subsection{Problem Statement}

\label{sec: problem_statement}
In subsequent sections, we refer to latent dimensions as `codes', and to the data generative factors as `factors'. Generative factors are those attributes that describe the perceptual differences between any two samples from dataset $\mathbf{X}$.


Consider a dataset $\mathbf{X} = {\mathbf{x}^{(i)}}_{i=1}^N$ comprising $N$ i.i.d. samples. We assume these samples $\mathbf{x}$ are generated by a random process $g: \mathbb{R}^k \rightarrow \mathcal{X}$, which takes the ground truth generative factors $\mathbf{z} \in \mathbb{R}^k$ as input and returns the generated data $\mathbf{x} \in \mathcal{X}$. We now consider a latent variable model capable of inferring the corresponding latent representation $\mathbf{c} \in \mathbb{R}^d$ of the data $\mathbf{x}$. This latent representation $\mathbf{c}$, analogous to $\mathbf{z}$, can be used to generate the corresponding data $\mathbf{x}$. The model simulates the random process of generating data $\mathbf{x}$ as follows: latent variables $\mathbf{c}$ are sampled from some prior distribution $p_\theta(\mathbf{c})$, and then the data $\mathbf{x}$ is sampled from a conditional distribution $p_\theta(\mathbf{x}|\mathbf{c})$. The model aims to approximate the desired data distribution $p_\theta(\mathbf{x}) = \int p_\theta(\mathbf{x}|\mathbf{c}) p_\theta(\mathbf{c}) d\mathbf{c}$.

Given the latent representations $\mathbf{c}$ learned by the trained latent variable model and the known ground truth generating factors $\mathbf{z}$,  we aim to obtain a method to quantitatively evaluate the disentanglement of the latent space $\mathbb{R}^d$ by giving a certain score $s \in \mathbb{R}$ according to the identified definitions of disentanglement.


\section{Existing Metrics and their Shortcomings}

In a recent survey, Carbonneau~\etal~\cite{carbonneau_metrics_2022} taxonomise the existing metrics into three categories viz. intervention-based, predictor-based and information-based. While this is a significant scholarly work, there appears to be a functional overlap between the intervention and predictor-based, as they both use either accuracy or weights from predictors to determine the factor-code relationships.

We take a more nuanced view of the metrics to highlight in depth the key differences in design, interpretation of disentanglement and thus investigate the metrics from a three-fold perspective, namely a) \textit{Aspect of measurement}, b) \textit{Detection of factor-code relationship} and c) \textit{Extent of characterisation}. We identify the good practices employed and the limitations of many of them~(cf.~\Cref{tab: metrics}). Recognising these weaknesses, we propose a new metric that categorically improves upon each~(cf.~\Cref{sec: edi}). Detailed mathematical formulations of the existing metrics consistent with this work are described in the appendix.
 
\subsubsection{Aspect of measurement.}

A close inspection of the metrics reveals a clear dichotomy in perspectives on disentanglement and consequently in the aspect of its measurement. Metrics that developed in studies with modularity as the key characteristic for disentanglement are designed to test if the factor is encoded by one or more codes, and tend to be calculated from the perspective of each code, 
On the other hand, metrics with compactness as the identified definition of disentanglement are designed to ensure that a code encodes only one factor at a time. These metrics tend to be calculated from the perspective of the factor.

The \emph{Modularity-centric} metrics include the BetaVAE metric, otherwise known as \emph{Z-diff}~\cite{higgins_beta-vae_2017}, and its successor, the FactorVAE metric or \emph{Z-min Variance}~\cite{kim_factorising_2019}. These early metrics are intervention-based \ie they use a predictor to determine which factor was fixed using statistics learnt from the latent codes.

The \emph{Compactness-centric} metrics include the \emph{Separated Attribute Predictability} metric (\emph{SAP})~\cite{kumar_sap_2018}, and \emph{Mutual Information Gap} (\emph{MIG})~\cite{chen_tcvae_2019}, followed by \emph{MIG-sup}~\cite{li_mig-sup_2020}, and \emph{DCIMIG}~\cite{sepliarskaia_how_2021} that were proposed to augment MIG with the ability to also capture modularity.


Other works propose to use a distinct metric to capture each aspect~\cite{ridgeway_fstatistic_2018}, including explicitness, separately. Eastwood~\etal~\cite{eastwood_dci_2022} continue in this vein and propose using three new metrics to compute modularity, compactness and explicitness, calling them disentanglement (D), completeness (C), and informativeness(I) under a unified framework entitled DCI. 


\subsubsection{Detection of relationship.}

The mechanism of detection of factor-code relationships varies across the metrics.

\emph{Prediction accuracy of classifiers:} 
The Z-diff and Z-min Variance metrics follow the intuition that code dimensions associated with a fixed factor should have the same value. So they fix one generative factor, while varying all the others, and use a linear classifier to predict the index of the fixed factor, based on the variance in each of the latent codes as in Z-diff or the index of the code with the lowest variance as input in Z-min Variance, such that the resulting classifier is a majority vote classifier. While this approach has the advantage of not making assumptions about factor-code relationships, these metrics require careful discretisation of the factor space (eg., the size and number of data subsets), and other design choices like classifier hyper-parameters and distance function. However, for random classifications, there is no code with the lowest variance, each code would get the same number of votes and so Z-min Variance would assign $\frac 1 d$ ($d$ being the number of latent codes) instead of $0$. 
The Explicitness metric in~\cite{ridgeway_fstatistic_2018} is measured similarly to Z-min Variance, with the difference that it uses the mean of one-vs-rest classification and ROC-AUC instead of accuracy. For discrete factors, SAP uses the classification accuracy of predicting factors using a classifier like Random Forest.

\emph{Linear correlation coefficient:} For continuous factors, SAP computes for each generative factor, the linear $R^2$ coefficient with each of the codes, then takes the difference between the largest and the second largest coefficient values to predict the code encoding it. This ensures that a large value is assigned when only one code is highly informative, and others negligible-- an intuition exploited by subsequent works like MIG, which employs mutual information instead of $R^2$. In the case of SAP, however, this limits the detectable factor-code relationship to a linear one.

\emph{Ad-hoc model:}  DCI utilises feature importance derived from classifiers. The authors~\cite{eastwood_dci_2022} originally proposed using a LASSO-based classifier with DCI to predict each generative factor from each latent factor and estimate scores from the weights and accuracy of the trained classifier. Hence the relationship matrix relies heavily on the ad-hoc model, requiring careful selection of the model and hyperparameters~\cite{carbonneau_metrics_2022}. Naturally, this metric thus may be prone to stochastic behaviour, which is less than ideal.

\emph{Mutual information (MI):} The use of mutual information to describe relationships was first proposed in MIG, and has since been adopted by many subsequent metrics, including Modularity score~\cite{ridgeway_fstatistic_2018}, MIG-sup, and DCIMIG. While this choice offers the advantage of not varying by implementation, and making no assumptions about the relationship between factors and codes, all these methods compute mutual information by binning and suffer from several challenges. We elaborate on this further in the next subsection on shortcomings.

\subsubsection{Extent of characterisation.} 
The ability of metrics to express the degree of modularity or compactness depends on the extent of characterisation.
The Z-diff metric uses maximum value to describe the extent of disentanglement. Consequently, it would not be capable of distinguishing whether a code captures primarily one factor or multiple factors. SAP and MIG take the difference between the top two entries to express the degree of completeness, which would not allow distinguishing whether a factor is encoded by two codes or by more than two codes. This yields limitations in functionality, discussed in the next subsection. MIG-sup, furthermore, is not affected by low information content, as it normalises mutual information by dividing by the entropy of the code, making it ignorant to information loss. DCI, in contrast, is designed well in this regard as it can express the degree of relationship by calculating $1 - entropy$ (where entropy is estimated from the probabilities derived from feature importance). Modularity is also equipped to express the degree well by calculating the deviation of all items from the maximum value.

\begin{table*}[!ht]
    \centering
    \caption{Summary of metrics in regards to measurement aspects viz. modularity (Mod), compactness (Comp) and explicitness (Expl), detection of relationship and extent of characterisation. Identified strengths and weaknesses in design are marked with +/- accordingly.}
    \label{tab: metrics}

    \begin{tabular}{lccc>{\centering\arraybackslash}p{4.5cm}>{\centering\arraybackslash}p{4.5cm}}
        \toprule
        Metric & Mod & Comp & Expl & Relationship Detection & Extent Characterisation \\
        \midrule
        Z-min Variance~\cite{kim_factorising_2019} & \checkmark & & & - Majority vote classifier accuracy & - Maximum value \\
        SAP~\cite{kumar_sap_2018} & & \checkmark & \checkmark & - Linear correlation (continuous); Predictive accuracy (categorical) & - Difference between top two \\
        MIG~\cite{chen_tcvae_2019} & & \checkmark & \checkmark & + Mutual information & - Difference between top two \\
        Modularity~\cite{ridgeway_fstatistic_2018} & \checkmark & & \checkmark & + Mutual information & + 1 - avg. squared deviations \\
        DCI~\cite{eastwood_dci_2022} & \checkmark & \checkmark & \checkmark & - Feature importance & + 1 - entropy \\
        MIG-sup~\cite{li_mig-sup_2020} & \checkmark & & & + Mutual information & - Difference between top two \\
        DCIMIG~\cite{sepliarskaia_how_2021} & \checkmark & & \checkmark & + Mutual information & - Difference between top two \\
        \bottomrule
    \end{tabular}
\end{table*}

\subsubsection{Shortcomings.}
\label{sec: functional_overview}

Abdi~\etal~\cite{abdi_preliminary_2019}, in a first attempt, reported inadequacies in the disentanglement metrics, noting discrepancies without delving into the underlying reasons. This observation spurred further investigations within the research community.  Chen~\etal~\cite{chen_tcvae_2019} examined metrics through the lens of robustness to hyperparameter selection during experiments and showed that the early modularity-centric metrics overestimated disentanglement. Sepliarskaia~\etal~\cite{sepliarskaia_how_2021}, in subsequent work, provided an initial theoretical analysis, unveiling specific cases of failures in the metrics, but lacked a controlled study. Carbonneau~\etal~\cite{carbonneau_metrics_2022} showed some controlled evidence of measurement of different properties and reported that the metrics differ in terms of measured properties and overall agreement. 
Surveying the literature, we identified the following major functional issues, that support our argument to have improved metrics:

a) \textit{Several metrics designed for a particular aspect fail in efficiently reflecting that aspect in all cases.} This is observed strongly in Z-diff and Z-min Variance which penalise modularity violations weakly~\cite{sepliarskaia_how_2021}. We conducted a systematic analysis to test metric calibration to confirm this and identify other discrepancies~(cf.~\Cref{sec:basic_tests}). This was also observed in the case of compactness-centric metrics like SAP and DCIMIG\footnote{When a factor is encoded by two codes, DCIMIG yields a score of $(\text{max} (I(c_0; z_0), I(c_1; z_0)) + I(c_2; z_1)) / (H(z_0) + H (z_1)) = 1$.}. In MIG, it was observed that it assigns a 0, when a factor is encoded by just two codes~\cite{carbonneau_metrics_2022}, indicating too strong a penalisation in partial entanglement.

b) \textit{Modularity-centric metrics are generally not equipped to capture compactness and disregard explicitness.} Since these metrics align $z_i$, with a corresponding set of codes, $c_i$, this strategy does not ensure that distinct codes are dedicated to unique factors. Further, they do not capture the extent of the factor-code relationship, and consequently cannot be reliably used to reflect disentanglement. 

c) \textit{Predictor-based methods can overfit and can be computationally expensive.} Metrics that use predictors to determine factor-code relationships can overfit when there are too few samples, resulting in overestimating explicitness~\cite{carbonneau_metrics_2022}. Furthermore, the complexity of the chosen model can result in undesirable computational complexity~(cf.~\Cref{sec:efficiency}).  

d) \textit{Existing information-based metrics are fraught with computational challenges.} The existing metrics that use mutual information using maximum likelihood estimators, that require quantisation of both spaces and parameterised sampling procedures. The existing formulations\footnote{it is commonly estimated as, $I(c, z) = \sum_{i=1}^{z} \sum_{j=1}^{c} P(i, j) \log \left( \frac{P(i, j)}{P(i) P(j)} \right)$.} expect a discretisation of spaces into bins, with the mutual information value estimation being sensitive to binning considerations. These pose further a challenge in scenarios dealing with high-dimensional or non-linear data~\cite{kraskov_mi_2004, carbonneau_metrics_2022}, discussed further in~\Cref{sec: non_linearity}. 

\section{Exclusivity Disentanglement Index (EDI)}
\label{sec: edi}

Having identified the best practices in design and their shortcomings, we exploit them to define the the disenglement aspects in a more intuitive and simple way, using the principle of exclusivity. 
In this section, we introduce our proposed metric EDI. First, we define impact intensity that measures the factor-code relationship. Next,
we define exclusivity which we subsequently use as the criteria to define and mathematically construct both modularity and compactness metric formulations.

\subsection{Impact Intensity} 

\label{sec: impact_intensity}
We measure the influence each of the factors $z_i$ have on the latent codes $c_i$ using a relationship matrix we call \textit{Impact Intensity}. We introduce two improvements in the computation of relationship matrix, namely, a) an improved estimator and b) no reliance on ad-hoc decision model.

As pointed out earlier, existing implementations of MI in metrics are unsuited to high-dimensional continuous variables and fraught with computational challenges~\cite{kraskov_mi_2004}. Naturally, a non-parametric estimator with no dependence on discretisation is more suitable.
A recently proposed method called MINE~\cite{belghazi_mine_2021} operates by training a small neural network to maximise a lower bound on the mutual information between two variables. As it involves no density estimation using maximum likelihood it is flexible and has been shown to converge to the true mutual information between high-dimensional variables~\cite{belghazi_mine_2021}. Linearly scalable in both dimensionality and sample size, it offers a significant advantage~(cf.~\Cref{sec: non_linearity,sec:efficiency}).

Thus, we propose computing the relationship matrix as follows: First, we calculate the following required variables: a) $I(\bm{c}_i; \bm{z}_j)$, signifying the mutual information computed between each factor $\bm{c}_i$ and each code $\bm{z}_j$; b) $I(\bm{c}_1, \bm{c}_2, \dots, \bm{c}_d; \bm{z}_j)$, signifying the mutual information between all codes $\bm{c}_1, \bm{c}_2, \dots, \bm{c}_d$ and each factor $\bm{z}_j$; and c) $H(\bm{z}_j)$, representing the entropy of each factor. We establish the relationship as $R(\bm{c}_i; \bm{z}_j) = \frac{I(\bm{c}_i; \bm{z}_j)}{I(\bm{c}_1, \bm{c}_2, \dots, \bm{c}_d; \bm{z}_j)}$, denoting the impact intensity of factor $\bm{z}_j$ on the code $\bm{c}_i$ among all codes. This, we argue, offers a more accurate representation of the relationship, as latent codes are learned from generative factors.

\subsection{Exclusivity}

The concept of exclusivity is crucial in both modularity and compactness. In modularity, we desire a code to capture a singular factor and exclude others. In compactness, it is expected that a factor is represented by a code without overlapping with others. This principle is fundamentally the inverse of impurity.

We propose an intuitive method to quantify the extent of exclusivity, which is defined as the difference between correctness (the maximum value) and incorrectness (the root mean square error of all other values). The objective is to maximise the difference between correctness and incorrectness.

Given a set of attributes $\{a_1, a_2, \ldots, a_n\}$, the exclusivity is mathematically represented as:

\[
\begin{aligned}
\text{Exclusivity}(a_1, a_2, \ldots, a_n) &= a_{i^*} - \sqrt{\frac{1}{n - 1} \sum_{\substack{i=1 \\ i \neq i^*}}^n a_i^2}, \\
&\quad \text{where} \quad i^* = \text{argmax}_i \, a_i.
\end{aligned}
\]

The aim is for the maximum value to be as high as possible, with the remainder as minimal as possible.

\subsection{Formulation}
By applying the aforementioned concepts of exclusivity to better depict the "extent", and impact intensity to capture factor-code relationships, we formulate the following metrics to measure modularity, compactness, and explicitness. 

\textbf{Modularity.} We formalize the metric for modularity, or disentanglement, of a latent code $\bm{c}_i$ as:
\[
D(\bm{c}_i) = \text{Exclusivity}\left(\bm{R}(\bm{c}_i; \bm{z}_1), \bm{R}(\bm{c}_i; \bm{z}_2), \ldots, \bm{R}(\bm{c}_i; \bm{z}_k)\right).
\]

The aggregate modularity score is then calculated as $D = \frac{1}{k} \sum_{i=1}^{d} D(\bm{c}_i)$, where $d$ denotes the code dimensionality, and $k$, representing the number of factors, signifies the maximum potential influence a single factor can exert. Notably, this framework may encounter complications due to correlated effects, wherein multiple codes capture a single factor. To address this challenge, we allocate to each code $\bm{c}_i$ and its predominantly associated factor $\bm{z}_{j^*}$ a score $S_{ij^*} = D(\bm{c}_i)$, while assigning $S_{ij} = 0$ for $j \neq j^*$. Accumulating these scores across all factors yields $S_j = \sum_i S_{ij}$ for each factor $j$. Ensuring that the score for each factor does not exceed $1$ (the maximum conceivable impact intensity for each factor is $1$), the final score is thus recalculated as $D = \frac{\sum_j \max(S_j, 1)}{k}$, facilitating an accurate assessment of modularity.

We then assign a score $S_{ij^*} = D(\bm{c}_i)$ to each code $\bm{c}_i$ and its most effective factor $\bm{z}_{j^*}$, and mark the others as $S_{ij} = 0$ for $j \neq j^*$. The overall disentanglement is finally calculated as:
\[ D = \frac {\sum_j{\min(S_j, 1) }} {k}, \textrm { where } S_j = \sum_i S_{ij}. \]

\textbf{Compactness.} The compactness of a generative factor $\bm{z}_j$ is calculated as:
\[
C(\bm{z}_j) = \text{Exclusivity}\left(\bm{R}(\bm{c}_1; \bm{z}_j), \bm{R}(\bm{c}_2; \bm{z}_j), \ldots, \bm{R}(\bm{c}_d; \bm{z}_j)\right).
\]
Accordingly, the overall compactness score, $C$, is determined by the average compactness across all factors:
\[
C = \frac{1}{k} \sum_{j=1}^{k} C(\bm{z}_j).
\]

\textbf{Explicitness.} For a generative factor $\bm{z}_j$, explicitness or informativeness is calculated as the ratio of the combined information content of the codes relative to $\bm{z}_j$ to the entropy of $\bm{z}_j$ itself:
\[
I(\bm{z}_j) = \frac{I(\bm{c}_1, \bm{c}_2, \ldots, \bm{c}_d; \bm{z}_j)}{H(\bm{z}_j)}.
\]
Hence, the aggregate measure of informativeness is the mean informativeness across all generative factors:
\[
I = \frac{1}{k} \sum_{j=1}^{k} I(\bm{z}_j).
\]

\section{Experiments}

\label{sec: experiments}
 In the following sections, we model the relation $c = \gamma(g(\bm z))$ as $\bm c = f(\bm z)$. Here, $f(\bm z)$ represents a fully-parameterised function controlling the factor-code relationship. For the experiments, factors $\bm z$ are sampled i.i.d from a discrete uniform distribution in~\Cref{sec:basic_tests,sec:efficiency}, and from a continuous uniform distribution $\mathcal{U} \in \{0,1\}$ in 
 \secrefrange{sec: non_linearity}{sec: decreasing_explicitness}. Following~\cite{carbonneau_metrics_2022}, we generate $N$ factors to form a set $Z$ and compute the corresponding set of codes $\bm c$ using the experiment-specific $f(\bm z)$ parameterised by $\alpha$, resulting in one representation. Unless otherwise specified, the factor and code dimensionality are kept equal ($k == d$). For each $\alpha$ within the chosen discrete range, we generate $M$ representations and aggregate over these for $s$ random seeds. In~\Cref{sec: practical_models}, representations are learnt using real latent variable models on a real-world dataset.

\subsection{Are the metrics well calibrated?}
\label{sec:basic_tests}
Motivated by discrepancies in our exploratory analysis, we first systematically assess metric behaviour via discrete boundary test cases for each of the aspects~\ie modularity, compactness, and explicitness.  Codes are arranged in that order with $1$ denoting a perfect aspect and $0$ completely imperfect. For example, $\#101$ indicates perfect disentanglement and explicitness, but imperfect compactness.


To form the factor space, we sample $N = 50,000$ points from a discrete uniform distribution with a one-to-one encoding. Each category is assigned a distinct code $(k = d = 2)$ unless: a) when modularity is low, we encode two factors into one code; b) when compactness is low, we encode a factor into two codes; or c) when informativeness is low, we randomly drop categories within the factors. We simulate a total of $2^3 = 8$ representative cases\footnote{detailed description in supplementary material}.  Results, reported in~\Cref{tab:basic_test} using $s = 50$ random seeds, confirm some intuitions, and previously reported observations while revealing interesting insights.

As discussed in~\Cref{sec: functional_overview}, not all metrics designed for specific aspects are well-calibrated. Z-min Variance, for instance, which is modularity-centric, fails to penalise modularity violations, with scores larger than 0.5 in low modularity scenarios~($\#000, \#001, \#010, \#011$). This stems from its assigning of the minimum score as $\frac{1}{d}$. The Modularity metric, while performing perfectly in high modularity cases, unexpectedly assigns high scores of 0.75 in low modularity scenarios too~($\#010$ and $\#011$). This is likely due to an error introduced by dividing the maximum term in the formula. DCI Mod correctly assigns low scores in low modularity cases of $\#000$ and $\#001$, however, it assigns relatively large scores of $>50\%$ in $\#010$ and $\#011$, indicating some influence of high compactness, which is not ideal. In contrast, EDI Mod assigns 0.43, reflecting low modularity relatively better.

The discrepancies appear in compactness-centric metrics as well. SAP, for instance, assigns a relatively low score of 0.33 in both high compactness scenarios ($\#010$ and $\#110$) but a higher score of 0.45 in the low compactness case of $\#101$, suggesting greater influence from other aspects. MIG also assigns relatively low scores of 0.41 and 0.45 in high compactness scenarios ($\#010$ and $\#110$), but a higher score of 0.49 in the less compact scenario of $\#101$. Its successor, MIG-sup, assigns a large score of 0.99 in both low~($\#100$, $\#101$) and high compactness scenarios~($\#110$, $\#111$) while tends to assign intermediate scores of about 0.5 in high compactness, low modularity scenarios ($\#010$). This shows a high influence from modularity but yields no clear interpretation of the captured aspects. Furthermore, the DCIMIG metric assigns a higher score to the low compactness case of $\#101$ confirming weak penalisation, as a consequence of two codes capturing different information extent about the factor. DCI Comp assigns very high scores to scenarios $\#100$ and $\#101$, which are highly modular despite low compactness. EDI Comp, in contrast, assigns lower scores. In terms of explicitness, EDI and DCI perform comparably. Overall the results indicate EDI to be better calibrated in comparison to the existing metrics.

\vspace{-0.1in}

\begin{table*}[ht]
    \centering
    \caption{Measurement results of all boundary test cases. Representative codes follow a (m,c,i) format with binary values indicating high or low. Results are reported as mean scores for 50 random seeds. Standard deviations are not included due to limited space, however, most values are close to 0.}
    \label{tab:basic_test}
    \small 
    \begin{tabularx}{\textwidth}{l  C C C C C C C C}
    \toprule
    \textbf{Nr.} & \textbf{000} & \textbf{001} & \textbf{010}  & \textbf{011}  & \textbf{100}  & \textbf{101}  & \textbf{110}  & \textbf{111}  \\
    \midrule
    Z-min Variance  & 0.57 & 0.55 & 0.62 & 0.67 & 1.00 & 1.00 & 1.00 & 1.00 \\
    \hline
    SAP          & 0.04 & 0.03 & 0.33 & 0.88 & 0.22 & 0.45 & 0.33 & 0.88 \\
    \hline
    MIG          & 0.06 & 0.034 & 0.41 & 0.82 & 0.23 & 0.49 & 0.45 & 0.99 \\
    \hline
    MIG-sup      & 0.11 & 0.03 & 0.54 & 0.63 & 0.99 & 0.99 & 0.99 & 1.00 \\
    \hline
    DCI Mod    & 0.08 & 0.00 & 0.57 & 0.57 & 0.99 & 1.00 & 0.99 & 1.00 \\
    DCI Comp   & 0.08 & 0.00 & 0.99 & 1.00 & 0.75 & 0.68 & 0.99 & 1.00 \\
    DCI Expl   & 0.44 & 1.00 & 0.44 & 1.00 & 0.44 & 1.00 & 0.44 & 1.00 \\
    \hline
    Modularity   & 0.25 & 0.25 & 0.75 & 0.75 & 1.00 & 1.00 & 1.00 & 1.00 \\
    \hline
    DCIMIG       & 0.05 & 0.02 & 0.17 & 0.46 & 0.38 & 0.75 & 0.46 & 1.00 \\
    \hline
    EDI Mod    & 0.11 & 0.02 & 0.43 & 0.43 & 0.99 & 0.99 & 0.99 & 0.99 \\
    EDI Comp     & 0.12 & 0.02 & 0.99 & 1.00 & 0.61 & 0.57 & 1.00 & 1.00 \\
    EDI Expl     & 0.45 & 0.99 & 0.45 & 0.99 & 0.45 & 0.99 & 0.45 & 0.99 \\
    \bottomrule
\end{tabularx}
\end{table*}

\subsection{How do the metrics deal with non-linearity?}
\label{sec: non_linearity}

\begin{figure*}[h]
    \centering
    \includegraphics[width = 0.9\linewidth]{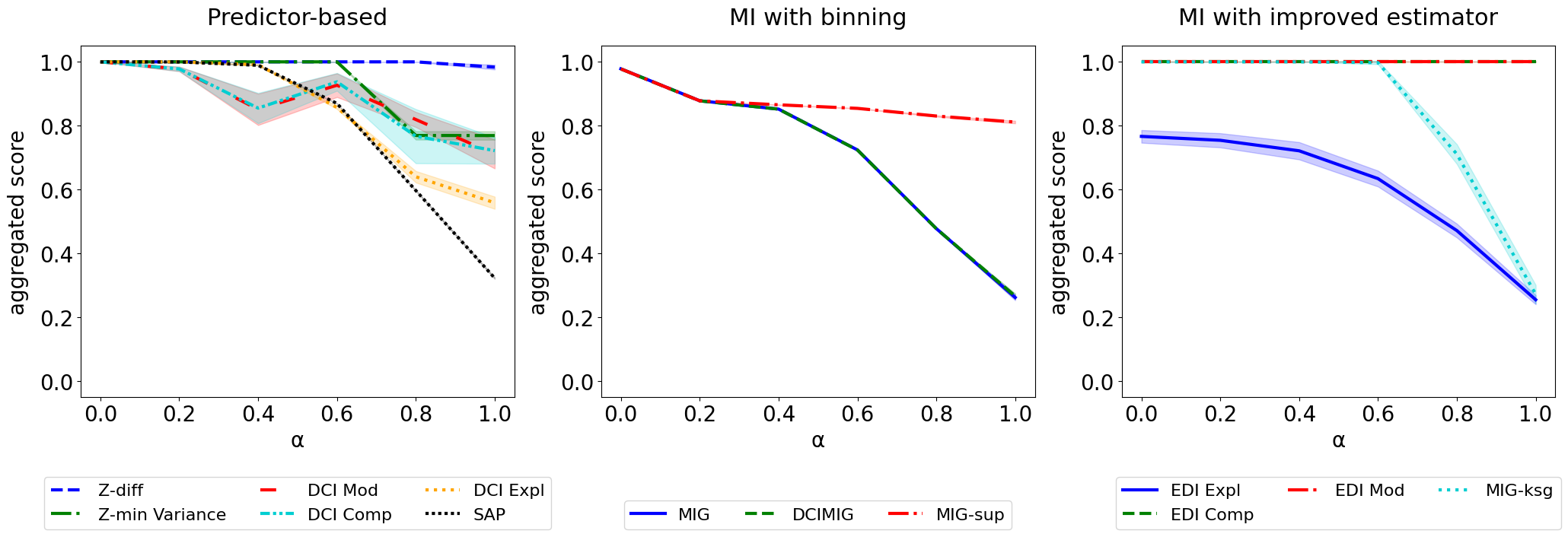}
    \caption{As $\alpha$ increases, the factor-code relationship becomes more non-linear. We see a decline in most metrics computing MI using binning, as well as metrics that use predictors. EDI, in comparison, exhibits good stability.}
    \label{fig:increasing_non_linearity}
\end{figure*}

The ability to attribute accurate scores when factor-code relationships are non-linear as in realistic data is a crucial property. A robust metric should exhibit negligible effect with increasing non-linearity. In this experiment, we simulate representations which are perfectly compact and modular, but the encoding function becomes increasingly non-linear. We use 
 $f(z)$ = \( 1000 - \alpha + 0.25 \tan\left(\omega(z - 0.5)\right) + 0.5\) where \(\omega = 2 \arctan\left(1000\alpha - \frac{0.25}{2}\right)\). As $\alpha$ increases, the curve becomes more steep but remains monotonic for \(z \in [0, 1]\). Using $k = d = 6$, we simulate $M = 50$ representations with $N = 20,000$ points sampled from $\mathcal{U} \in [0,1]$. For $s = 50$ seeds, we report the aggregated scores in~\Cref{fig:increasing_non_linearity}. 
 
This experiment perfectly challenges the complexity of the predictors employed by the metrics, and highlights the potential issues inherent to mutual information computation using density estimation, yielding interesting insights. Metrics that calculate MI using binning methods generally perform inadequately. To illustrate this, we contrasted MIG to its alternative variant implemented with a non-parametric estimator, KSG~\cite{kraskov_mi_2004}. MIG-ksg demonstrates greater stability until reaching $\alpha = 0.6$, after which it gradually declines. Metrics using linear models like SAP, exhibit instability as non-linearity increases. This is also observed in DCI, which in its original implementation uses a LASSO classifier. Both DCI Mod and Comp decline and become more variable as non-linearity increases. In contrast, Z-diff and Modularity scores exhibit stability throughout the experiment. EDI Mod and Comp consistently assign a perfect score of $1$ throughout too, indicating robustness in this setting. For explicitness, a slight reduction in mutual information is expected.

\subsection{How do the metrics behave on decreasing disentanglement?}

\label{sec: decreasing_disentanglement}

Next, we evaluate the performance of the metrics as a perfectly disentangled representation gradually transitions to an entangled state. We conduct an experiment where we linearly reduce the modularity and compactness of the representation while maintaining explicitness.  To describe the factor-code relationship, we employ $f(\bm z) = \bm {zR}$, with 
$$
\small 
\bm R = 
\begin{pmatrix}
1 - \alpha & \alpha & 0 & \cdots & 0 \\
0 & 1 - \alpha & \alpha & \cdots & 0 \\
\vdots & \ddots & \ddots & \ddots & \vdots \\
0 & \cdots & 0 & 1 - \alpha & \alpha \\
\alpha & 0 & \cdots & 0 & 1 - \alpha
\end{pmatrix}.
$$



\begin{figure*}[h]
    \centering
    \includegraphics[width = 0.9\linewidth]{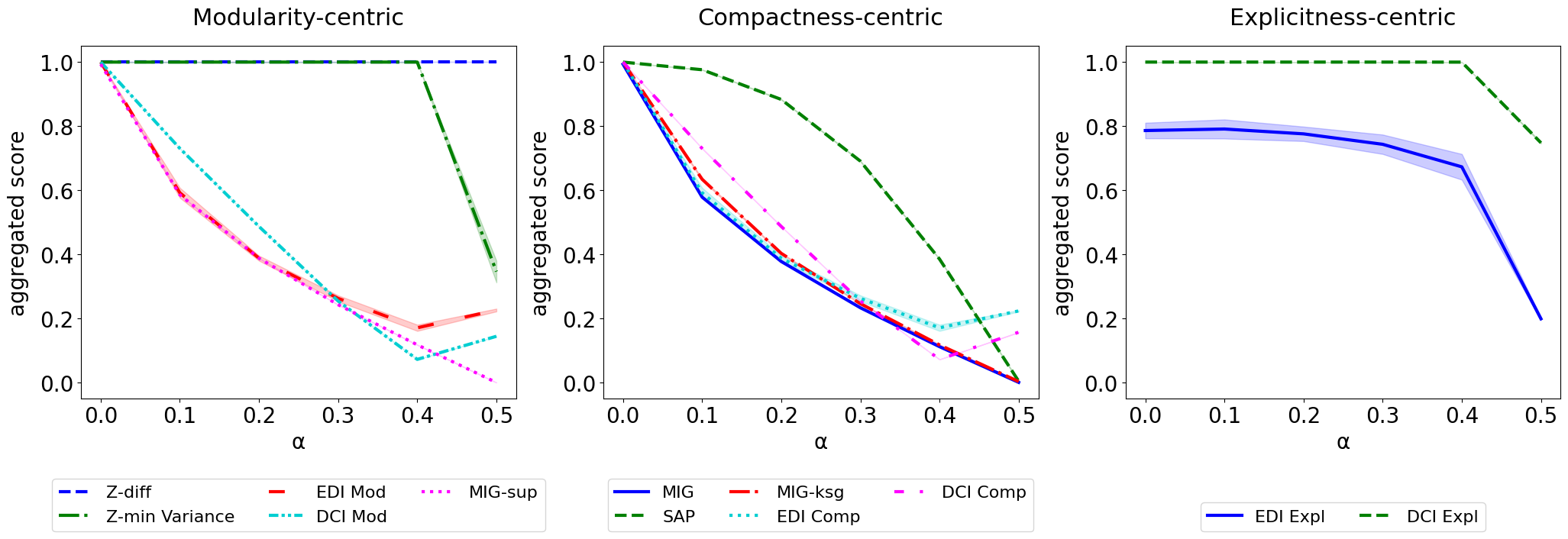}
    \caption{As $\alpha$ increases, the representation becomes less modular and compact. EDI and DCI perform adequately, whereas MIG, SAP assign $0$ with partial entanglement and Z-diff, Z-min Variance fail to observe any difference. }
    \label{fig:decreasing_disentanglement}
\end{figure*}

Like in the previous experiment, we use $k = d = 6$, and simulate $M = 50$ representations with $N = 20,000$ points. For $s = 50$ seeds, we report the aggregated scores in ~\Cref{fig:decreasing_disentanglement}. As the parameter $\alpha$ increases, we expect a linear decrease in all metrics dealing with modularity or compactness, though not reaching $0$ entirely\footnote{since increasing $\alpha$ does not lead to perfect entanglement, \ie a factor equally represented by all codes. Instead, only two codes capture a factor.}. Z-diff and Z-min Variance metrics fail completely in this regard. Conversely, both modularity and compactness components of EDI and DCI respectively demonstrate robust performance. DCI Expl, which does not represent true mutual information remains largely unaffected. It is also prone to overfitting and hence may overestimate explicitness~\cite{carbonneau_metrics_2022}. In comparison, EDI Expl exhibits a drop when one factor becomes equally represented by two codes. Most information-based metrics also perform well, however, assign zero value already when only one factor or code becomes fully entangled. 

\subsection{How do the metrics deal with noise?}

\label{sec: decreasing_explicitness}

\begin{figure*}[h]
    \centering
    \includegraphics[width = 0.9\linewidth]{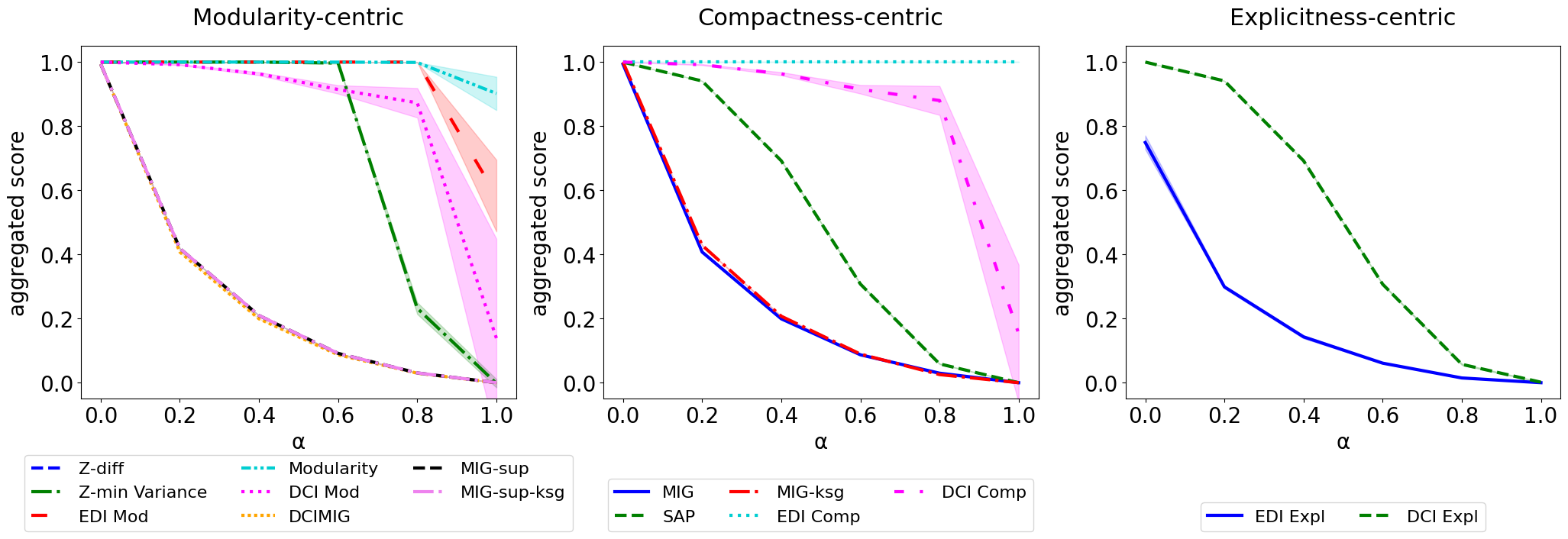}
    \caption{As $\alpha$ increases the representation becomes more noisy. We expect explicitness measuring metrics to gradually reach 0, but modularity and compactness metrics should stay unaffected. EDI exhibits greater stability here in comparison to others.}
    \label{fig:decreasing_explicitness}
\end{figure*}

In this segment, we investigate how the metrics behave when we keep modularity and compactness intact, but gradually reduce explicitness. Choosing $ f(z) = (1- \alpha)z + \alpha n$, and keeping the setting consistent as before, we report the results in~\Cref{fig:decreasing_explicitness}.
In this simulation, we expect the metrics representing explicitness to decrease gradually. In this regard, both EDI Expl and DCI Expl perform adequately, however unlike DCI, EDI Expl does not assign a perfect score of 1 here due to the true mutual information being less than 1.
Metrics representing modularity and compactness should exhibit unchanged behaviour under noise. Here we see a larger contrast between the metrics. While MIG, SAP, and Modularity metric decrease gradually and reach 0, Z-min Variance collapses rapidly after the middle mark. DCI Mod and Comp also decrease, first slowly, then quite rapidly as $\alpha$ approaches $0.8$. Here we can see strikingly more stability in EDI Mod and Comp. In fact, even Z-diff appears to be robust here.

\subsection{How do the metrics compare on resource efficiency? }
\label{sec:efficiency}

\begin{figure}[h]
    \centering
    \includegraphics[width = 0.55\textwidth, height = 4.5 cm]{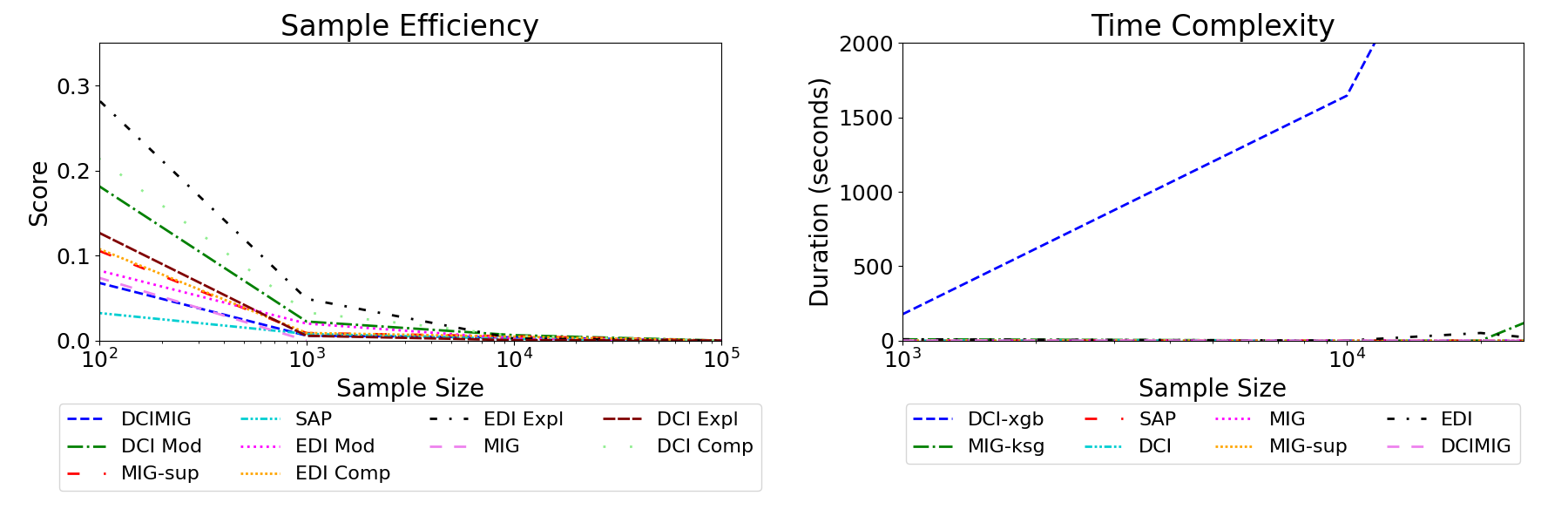}
    \caption{Comparing sample efficiency (left) and time complexity (right). A metric is more sample efficient if it shows a smaller difference in its score as sample size increases. Time complexity is assessed by examining the rate of change in computation duration as the sample size increases. Here we see a clear downside of using complex predictors to model factor-code relationships.}
    \label{fig:efficiency_complexity}
\end{figure}
Here, we evaluate and compare the metrics in terms of sample efficiency and time complexity. To test sample efficiency, we compute the difference in estimated scores when using subsets of data $N \in \{100,1000,10000,10000\}$ against the full sample size of $N = 100,000$. We keep the experimental setup as in~\cref{sec:basic_tests}, with a difference that we use only 10 random seeds, and report the mean differences in~\Cref{fig:efficiency_complexity} (left). 
We observe the minimum samples required to reliably estimate scores vary across metrics, as a result of design choices. While most metrics converge around the 10,000 sample mark, it becomes evident that classifiers-based metrics such as DCI necessitate larger sample sizes for optimal performance, whereas metrics reliant on MI require fewer samples. In this regard, EDI generally is more sample-efficient than DCI, with the exception of its explicitness component, which needs more samples to reliably compute mutual information. 

In terms of time complexity, most metrics are constant or (sub)linear. We observed EDI to be linear, and for DCI, it depends on the complexity of the ad-hoc model. If one were to choose complex models like random forest or XGBoost (DCI-xgb) to model non-linearity better as recommended~\cite{carbonneau_metrics_2022}, this would come with a serious disadvantage of the curse of dimensionality~(cf.~\Cref{fig:efficiency_complexity} (right)).



\subsection{Metric Agreement on Real Dataset}
\label{sec: practical_models}

\begin{figure}[!ht]
  \centering
  \label{fig: corr}
  \includegraphics[width=0.45\textwidth]{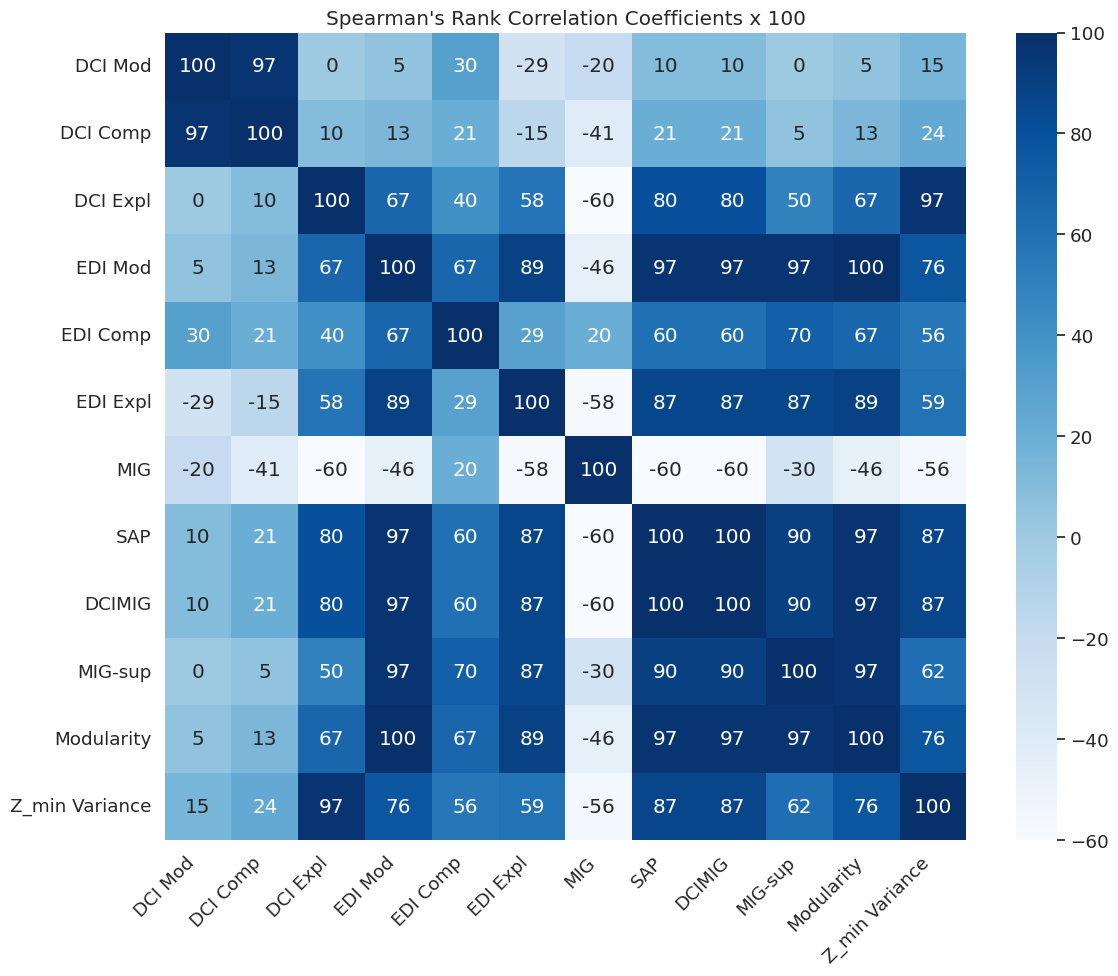} 
  \caption{Metric correlations on Shapes3D using Spearman's rho.}
\end{figure}

It was observed in previous works that metrics do not correlate on complex datasets, and the correlations may not be consistent across datasets~\cite{locatello_challenging_2019}. While we do not test consistency in this regard, we test general agreement of the metrics on a popular dataset used in the domain, namely Shapes3D\footnote{https://github.com/google-deepmind/3d-shapes}, in order to test if EDI can be applied in real settings. We heuristically opted to utilise FactorVAE~\cite{kim_factorising_2019} and BetaVAE~\cite{higgins_beta-vae_2017} for learning representations. For FactorVAE, we chose  $\gamma \in \{2, 4, 6, 8, 10\}$, and for BetaVAE, $\beta \in \{2, 3, 4, 5\}$. For $5$ random seeds, this resulted in $45$ representations in total. Next, we produce a ranking of the learned representations on the scores and calculate the agreement between the rankings for each pair of metrics using Spearman's coefficient~(cf.~\Cref{fig: corr}). 

We observe EDI to display strong correlations with SAP, DCIMIG, MIG-sup, Z-min Variance and perfect correlation with Modularity, indicating general agreement on both modularity and compactness aspects. The exception in this case is DCI which does not appear to correlate with most metrics. It might be that DCI required more hyperparameter tuning. MIG demonstrates a negative correlation with all metrics except EDI compactness. indicating that both measure similar properties to an extent.

\section{Conclusion}
In this study, we conducted a comprehensive analysis of existing metrics for evaluating disentanglement, elucidating differences in their assumptions, design, and functionality. By focusing on best practices, we formulated a novel metric, EDI, grounded in the intuitive and novel concept of exclusivity. Through controlled simulations, we demonstrated EDI to be well-calibrated, and better in comparison to existing metrics on non-linearity, resource efficiency and robustness under noise. These observations indicate a better suitability of EDI in supervised disentanglement measurement. However, it is essential to acknowledge that several pertinent questions remain open. Specifically, the development of unsupervised metrics has not progressed well, which has restricted the evaluation of disentangled representations in real-world scenarios. We hope and aim for further research in this direction to address this gap, as it holds promise for enhancing the practical utility of disentanglement evaluation in diverse contexts.

\printbibliography

\appendix


\subsection{Existing Metric Formulations}
\label{appendix: metrics}
\subsubsection{Z-diff (BetaVAE) Metric}
Higgins~\etal~\cite{higgins_beta-vae_2017} introduced the BetaVAE based on the notion that dimensions capturing the constant generative factor should match, while others vary. This metric aims to capture modularity by computing the following steps:
\begin{enumerate}[label=(\alph*)]
\item Selecting a generative factor $z_{k}$.
\item Choosing a pair of samples, $s_{1}$ and $s_{2}$, with $z_{k}$ constant while other factors vary.
\item Generating latent codes $c_{1}$ and $c_{2}$.
\item Calculating pairwise distortion:
\begin{equation}
e = (|c_{1,i} - c_{2,i}|), 1 \leq i \leq |z|
\end{equation}
\item Repeating the above steps to train a linear classifier predicting the fixed generative factor, with Z-diff indicating classifier precision.
\end{enumerate}

\subsubsection{Z-min Variance (FactorVAE) Metric}
Kim~\etal~\cite{kim_factorising_2019} proposed a metric similar to Z-diff, based on the assumption that latent codes capturing a constant generative factor should remain consistent. The method normalises each latent code by its dataset-wide standard deviation. The latent dimension with the least variance and the index of the constant factor form a sample for a linear classifier, assessing the classifier's precision.

\subsubsection{Separated Attribute Predictability (SAP)}
Kumar~\etal~\cite{kumar_sap_2018} developed the SAP metric, based on a matrix of informativeness $I$, with each entry $I_{i,k}$ representing a linear regression from latent code $c_{i}$ to generative factor $z_{k}$. The SAP score is:
\begin{equation}
SAP(c,z) = \frac{1}{D} \sum_{j} \left(I_{i_{k},k} - \max_{l \neq i_{k}} I_{l,k}\right), ; i_{k} = \argmax_{i} I_{i,k}
\end{equation}

\subsubsection{Modularity Score}
Ridgeway~\etal~\cite{ridgeway_fstatistic_2018} proposed a modularity metric of a latent cod $c_{i}$ as: 
\begin{equation}
    modularity = 1 - \frac{ \sum_{k \in \Omega_{\neq *}} I(k,c_{k})}{I(z_{*},c_{k})^{2} \times (M-1)},
\end{equation}
where $z_{*}$ represents the factor that has the highest mutual information, $\Omega_{\neq *}$ denotes the set of all the generative factors except $z_{*}$, and $M$ represents the number of factors.

\subsubsection{MIG}
The Mutual Information Gap (MIG), as detailed by Chen~\etal~\cite{chen_tcvae_2019}, estimates disentanglement through the empirical mutual information between latent codes and generative factors:

$$ \frac 1 K \sum_{k} \frac 1 {H(z_j)} \left ( I(c_{i^*}; z_j ) - \max_{i \neq i^*} I (c_i; z_j) \right ) \text{ ,} $$
where $i^* = \textrm{argmax}_i I(c_i; z_j) $, $\textrm{H}(z_j)$ is the entropy of $z_j$.

\subsubsection{MIG-sup}
As a complement to MIG, MIG-sup, introduced by Li~\etal~\cite{li_mig-sup_2020}, addresses MIG's limitation regarding modularity. It averages differences between the top two mutual information values for each code and factor.

\subsubsection{DCI}
The idea behind DCI~(\cite{eastwood_dci_2022}) is that it is possible to recover generative factors from latent units. Therefore, in order to compute \emph{disentanglement}, \emph{completeness} and \emph{informativeness}, a model $M$ trained to reconstruct generative factors from latent units is needed. The sub-model for predicting the generative factor $z_j$ from latent codes $c$ should be able to calculate the feature importance of each input latent code unit $c_i$ and the feature importance is denoted as $R_{ij}$.

\subsubsection*{Disentanglement}
The ``probability'' that $c_i$ being important for predicting $z_j$ in all factors is simulated as $P_{ij} =  \frac {R_{ij}} {\sum_k R_{ik}} $. The disentanglement score for the code $c_i$ is then calculated as:
$$D_i = 1 - \textrm{H}(P_{c_i}) \text{, where } \textrm{H}(P_{c_i}) = -\sum_{k} P_{ik} \log_K P_{ik}$$

\subsubsection*{Completeness}
Similarly, the ``probability'' that $c_i$ is important in all codes for predicting $z_j$ is $P_{ij} = R_{ij}$. The completeness score for the generative factor $z_j$ is then calculated as:
 $$C_j = 1 - H(P_{z_j}) \text{, where } H(P_{z_j}) = -\sum_{d} P_{dj} \log_D P_{dj}$$

\subsubsection*{Informativeness}
The informativeness of the generative factor $z_j$ is estimated as the prediction error of $z_j$ from the latent codes $c$.
$$I_j = \textrm{Error}(z_j, \hat z_j) =  \textrm{Error}(z_j, M_j(c)) \text{}$$

\subsubsection{DCIMIG}
DCIMIG or 3CharM claim to satisfy the three characters of disentanglement simultaneously (\cite{sepliarskaia_how_2021}). It is calculated as follows:
\begin{enumerate}[label=(\alph*)]
    \item calculate the disentanglement score for each latent code $c_i$ as $D(c_i) = I(c_i; z_{j^*} ) - \max_{j \neq j^*} I (c_i; z_j) $, where $j^* = \textrm{argmax}_j I(c_i; z_j) $.
    \item calculate the disentanglement score for each generative factor $z_j$ as \\ $D(z_j) = \max_i D(c_i)$, where $j = j^*$ in calculating $D(c_i)$. That is, $D(z_j)$ is maximum value among the disentanglement scores of the codes that capture $z_j$. If no code capture $z_j$, $D(z_j) = 0$.
    \item 3CharM is then defined as $\frac { \sum_j D(z_j)} { \sum_j H(z_j)}$
\end{enumerate}

\section{Experiment Setup for Validation Disentanglement Metrics on Practical Models}

\subsection{Data}
\textbf{Shapes3D}

Shapes3D~\footnote{https://github.com/google-deepmind/3d-shapes}is a dataset of 3D shapes procedurally generated from 6 ground truth independent latent factors. These factors are:

\begin{itemize}
  \item Floor (colour) hue: 10 values linearly spaced in $[0, 1]$
  \item Wall (colour) hue: 10 values linearly spaced in $[0, 1]$
  \item Object (colour) hue: 10 values linearly spaced in $[0, 1]$
  \item Scale: 8 values linearly spaced in $[0, 1]$
  \item Shape: 4 values in $[0, 1, 2, 3]$
  \item Orientation: 15 values linearly spaced in $[-30, 30]$
\end{itemize}

All possible combinations of these latents are present exactly once, generating $N = 480,000$ total images. All factors are sampled uniformly and independently of each other. 

\subsection{Model}

We select latent variable models that enforce disentanglement by regularizing the encoding distribution $q_\phi(z|x)$ in the VAE. Theoretically, latent representations learned by the selected model should have better disentanglement than those learned by VAE. We select BetaVAE and FactorVAE for experimentation. For a fair comparison, we applied a common encoder/decoder architecture for all VAE variants, as described in \Cref{tab:model}. For the discriminator in FactorVAE, we used the same model architecture as in FactorVAE: a feed forward neural network that has six hidden layers with 1000 neurons each, using a leaky ReLU of factor 0.2 as activation, and an output layer with two output units.

\begin{table}
    \centering
    \caption{Details of encoder and decoder architecture in the experiments}
    \label{tab:model}
    \begin{tabular}{l l}
        \toprule
        Encoder                                       & Decoder                                                \\
        \hline
        Input: $ n_i \times 64 \times 64$             
        & Input: $n_c$ \\
        Conv: $32 \times 4 \times 4$ (stride 2), ReLU
        & Linear: 256, ReLU\\
        Conv: $32 \times 4 \times 4$ (stride 2), ReLU 
        & Linear: 1024, ReLU \\
        Conv: $64 \times 4 \times 4$ (stride 2), ReLU
        & ConvTranspose: $64 \times 4 \times 4$ (stride 2), ReLU \\
        Conv: $64 \times 4 \times 4$ (stride 2), ReLU
        & ConvTranspose: $64 \times 4 \times 4$ (stride 2), ReLU \\
        Linear: 256, ReLU
        & ConvTranspose: $32 \times 4 \times 4$ (stride 2), ReLU \\
        Linear: $2\times n_c$ 
        & ConvTranspose: $32 \times 4 \times 4$ (stride 2) \\
        \bottomrule
    \end{tabular}
\end{table}



\subsection{Training}

Again for better comparison, we fixed all the training hyperparameters used to train VAE, as detailed in \Cref{tab:hyperparameters}. All parameters are set as closely as possible to previous works~\cite{locatello_challenging_2019,kim_factorising_2019}, while also taking into account actual training speed and performance as much as possible. In addition, we use Adam optimizer with learning rate 1e-4, $\beta_1=0.5$, $\beta_2=0.9$ for training the discriminator of FactorVAE.

\begin{table}[h]
    \centering
    \caption{Training hyper-parameters}
    \label{tab:hyperparameters}
    \begin{tabular}{l l}
        \toprule
        Parameter Key         & Value                        \\
        \midrule
        training epochs & 128                            \\
        batch size      & 64                             \\
        optimizer       & Adam: $\beta_1$ 0.9, $\beta_2$ 0.999 \\
        learning rate   & 1e-4                           \\
        reconstruction loss  & binary cross entropy \\
        \bottomrule
    \end{tabular}
\end{table}

Of the data, $ 90\% $ is used for training and the remaining $ 10\% $ is used for testing. Considering the robustness, we train each model with 10 different random seeds. The metrics will examine the representation learned by each model, and finally we aggregate the evaluation results.


\begin{table}[!ht]
    \centering
    \small
    \caption{The basic test cases used in the experiment on calibration, where columns Mod, Comp and Expl indicate the low or high in modularity, compactness, and explicitness, respectively.}\label{tab:cases}

    \begin{tabular}{c c c c c l}
        \toprule
        Mod & Comp & Expl & factors & codes & description \\
        \midrule
        0 & 0 & 0 & $z_1z_2$ & $c_1c_2$ & same as $\#001$, with reduced information \\
        0 & 0 & 1 & $z_1z_2$ & $c_1c_2$ & $c_1c_2$ together encode $z$ \\
        0 & 1 & 0 & $z_1z_2z_3$ & $c_1c_2$ & same as $\#011$, with reduced information \\
        0 & 1 & 1 & $z_1z_2z_3$ & $c_1c_2$ & $c_1$ encodes $z_1z_2$, $c_2$ encodes $z_3$ \\
        1 & 0 & 0 & $z_1z_2$ & $c_1c_2c_3$ & same as $\#101$, with reduced information \\
        1 & 0 & 1 & $z_1z_2$ & $c_1c_2c_3$ & $c_1c_2$ together encode $z_1$, $c_3$ encodes $z_2$ \\ 
        1 & 1 & 0 & $z_1z_2$ & $c_1c_2$ & same as $\#111$, with reduced information \\
        1 & 1 & 1 & $z_1z_2$ & $c_1c_2$ & $c_1$ encodes $z_1$, $c_2$ encodes $z_2$ \\
        \bottomrule
    \end{tabular}
\end{table}
\end{document}

